\documentclass[journal]{IEEEtran}

\usepackage{hyperref}
\usepackage{booktabs}
\usepackage{tabularx}
\usepackage{algorithm}
\usepackage{algorithmicx}
\usepackage{algpseudocode}  
\usepackage{colortbl}
\usepackage{epsfig}
\usepackage{graphicx}
\usepackage{amsmath}
\usepackage{amssymb}
\usepackage{enumitem}
\usepackage{xcolor}
\usepackage{xspace}
\usepackage{balance}
\usepackage{multirow}
\usepackage{mathtools}
\usepackage{rotating}
\usepackage{url}
\usepackage{tabulary}
\usepackage{array}
\usepackage[export]{adjustbox}
\usepackage{ctable}
\usepackage{subcaption}
\usepackage{hhline}

\def\Vec#1{{\boldsymbol{#1}}}

\definecolor{myblue1}{RGB}{0,127,255}
\def\CellB#1{\cellcolor{myblue1!25}#1}
\definecolor{mygreen1}{RGB}{128,255,0}
\def\CellG#1{\cellcolor{mygreen1!25}#1}
\definecolor{myyellow1}{RGB}{255,255,102}
\def\CellY#1{\cellcolor{myyellow1!25}#1}

\newcommand{\fig}{{Figure}\@\xspace}
\newcommand{\tab}{{Table}\@\xspace}
\newcommand{\eqn}{{Eq.}\@\xspace}

\newcommand{\ie}{{i.e.}\@\xspace}
\newcommand{\eg}{{e.g.}\@\xspace}
\newcommand{\etal}{{\it et~al.}\@\xspace}

\graphicspath{{./src_figs/}}

\begin{document}
\title{Interact as You Intend: Intention-Driven Human-Object Interaction Detection}

\author{
	Bingjie~Xu,
	Junnan~Li,
	Yongkang~Wong,~\IEEEmembership{Member,~IEEE,}
	Qi~Zhao,~\IEEEmembership{Member,~IEEE,}
	and Mohan~S.~Kankanhalli,~\IEEEmembership{Fellow,~IEEE}
	\IEEEcompsocitemizethanks{
		Bingjie~Xu and Junnan~Li are with the Graduate School for Integrative Sciences and Engineering, 
		National University of Singapore (email: \{bingjiexu, lijunnan\}@u.nus.edu).
		Yongkang~Wong and Mohan~S.~Kankanhalli are with the School of Computing, 
		National University of Singapore (email: yongkang.wong@nus.edu.sg and mohan@comp.nus.edu.sg). 
		Qi~Zhao is with the Department of Computer Science and Engineering,
		University of Minnesota (email: qzhao@cs.umn.edu).		
	}
	\thanks
	{	
	This research is supported by the National Research Foundation, Prime Minister's Office, Singapore under its Strategic Capability Research Centres Funding Initiative.	
	Manuscript received 20 July, 2018; revised XXX, 201X.
	}
}

\markboth{A SUBMISSION TO IEEE TRANSACTIONS ON MULTIMEDIA}%
{Shell \MakeLowercase{\textit{et al.}}: Bare Demo of IEEEtran.cls for IEEE Journals}
 
\maketitle
\begin{abstract}
	
The recent advances in instance-level detection tasks lay strong foundation for genuine comprehension of the visual scenes.
However, the ability to fully comprehend a social scene is still in its preliminary stage. 
In this work, 
we focus on detecting human-object interactions (HOIs) in social scene images, 
which is demanding in terms of research and increasingly useful for practical applications.
To undertake social tasks interacting with objects, 
humans direct their attention and move their body based on their intention.
Based on this observation,
we provide a unique computational perspective to explore human intention in HOI detection.
Specifically,
the proposed human intention-driven HOI detection (iHOI) framework models human pose with the relative distances from body joints to the object instances.
It also utilizes human gaze to guide the attended contextual regions in a weakly-supervised setting. 
In addition, 
we propose a hard negative sampling strategy to address the problem of mis-grouping.
We perform extensive experiments on two benchmark datasets, namely V-COCO and HICO-DET.
The efficacy of each proposed component has also been validated.

\end{abstract}


\IEEEpeerreviewmaketitle

\section{Introduction}
\label{sec:introduction}

In recent years, 
computer vision models have made tremendous improvements, especially in the instance-level tasks such as image classification and object detection~\cite{he2016deep,ren2015faster,lin2017feature}.
The advances in these fundamental tasks bear great potential for many fields,
including security, medical care and robotics~\cite{Chu_2018_TMM,wang2015atypical,Hayes_2017_ICRA}.
Enabling such applications requires deeper understanding of the scene semantics beyond instance-level understanding.
Existing efforts on the high-level semantic understanding include visual relationships inference~\cite{lu2016visual,li2017dual}, 
scene graphs generation~\cite{li2017scene}, 
and visual reasoning~\cite{Johnson_ICCV_2017}. 
In this work, 
we focus on an important task that is human-centric, 
namely human-object interaction (HOI) detection, 
stepping towards higher level scene understanding.

\begin{figure}[!t]
	\centering
	\includegraphics[width=1.0\linewidth]{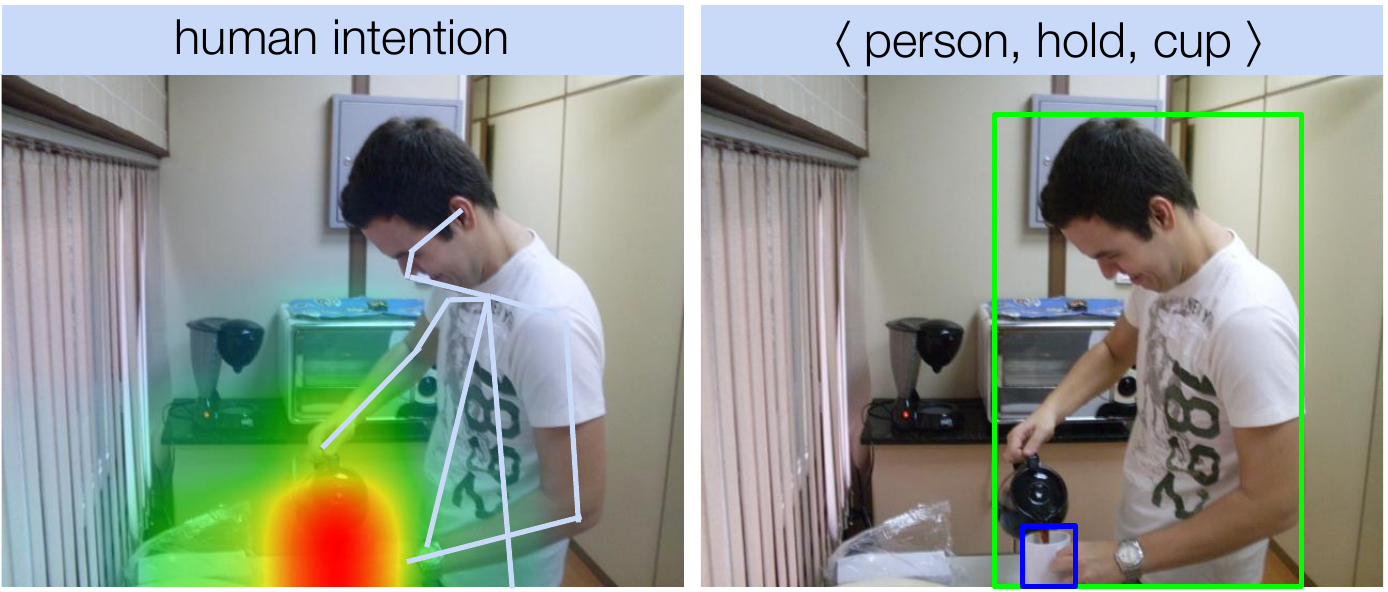}
	\caption
		{
		An example where the actor's intention is informative of the HOI {\small $\langle \mathsf{person}, \mathsf{hold}, \mathsf{cup} \rangle$}.
		The intention is represented using the attended regions and body pose.
		Specifically, he is fixating at the regions around the cup that he is interacting with,
		and his posture implicitly conveys his intention.
		} 
		\vspace{-1ex}
\label{fig:intro}
\end{figure}

The task of HOI understanding~\cite{gupta2009observing, Wang_2017_TMM, gkioxari2017detecting, gupta2015visual, chao2017learning} is formulated as identifying the {\small $\langle \mathsf{human}, \mathsf{action}, \mathsf{object} \rangle$} triplets.
It is a facet of visual relationships critically driven by humans.
In contrast to general visual relationships involving verbs, prepositional, spatial, and comparative phrases, 
HOI understanding focuses on direct interactions (actions) performed on objects (\eg~{\it a person is holding a cup} in \fig~\ref{fig:intro}).
Precise detection and inference of HOIs are increasingly needed in practical applications, 
such as 
development of collaborative robotics, activity mining in social networks, and event detection in surveillance~\cite{Hayes_2017_ICRA,Yang_2018_TMM,Chu_2018_TMM}.
Nevertheless,
it still remains a challenging research problem due to the fine granularity of actions and objects in social scenes.
Earlier approaches for HOI detection mainly focus on the representation of visual data,
such as joint modeling of body poses, spatial configuration and functional compatibility in images~\cite{gupta2009observing,yao2012recognizing, hu2013recognising}.  
In recent years, 
several large-scale datasets with diverse interaction classes have enabled fine-grained exploration of HOIs~\cite{gupta2015visual,chao2017learning,chao2015hico,zhuang2017care}. 
Motivated by advances in deep learning, 
especially the success of Convolutional Neural Network in object detection and classification,
recent works utilize those datasets to learn deep visual representations of human and object for HOI detection~\cite{gkioxari2017detecting,gupta2015visual,chao2017learning,Gao_BMVC_2018}.
However, those works do not take special consideration that a human often exhibits purposeful behaviors with intention in mind to complete tasks.
For example, in \fig~\ref{fig:intro}, the person is lifting the kettle and holding the cup, gazing around the target cup -- intended to pour water into the cup. 

In cognitive studies,
human intention is reported to commonly unveil complementary cues to explain the behaviors of individuals attempting to accomplish certain tasks~\cite{malle1997folk},
driving the coordination of eye and body movements~\cite{pelz2001coordination}.
For example,
when interacting with a specific object,
human tends to exhibit corresponding intention by adjusting position, pose and shifting attention (see \fig~\ref{fig:intro}).
By perceiving the latent goal, we can facilitate the inference of the interactions.
In this work,
we provide a novel computational perspective to exploit two forms of human intention that is visually observable:
1) human gaze, which explicitly conveys intention;
2) human body posture, which implicitly conveys the intention. 
The work most related to ours is the one characterizing human intention with attention and body skeleton~\cite{Wei_CVPR_2018}.
Nevertheless, 
human intention has yet been investigated in the context of HOI detection in an integrative manner.
Also, we offer more robustness to inaccurate gaze localization by exploring multiple contextual regions driven by gaze.

We utilize gaze to guide the model in exploring informative object instances in a scene.
The scene information has exhibited positive influence in various recognition tasks~\cite{li2017dual,Gao_BMVC_2018, zhou2017places,gkioxari2015contextual}.
One direction to utilize this information is directly extracting the visual representations from Scene-CNN~\cite{Gao_BMVC_2018, zhou2017places}.
However,
it is inefficient in some tasks due to the lack of explicit instance-level information.
Another direction is to leverage the visual cues from surrounding objects~\cite{li2017dual,gkioxari2015contextual}.
Such cues could be informative as a semantic scene constraint.
To leverage the informative scene instances, 
the existing approaches learn from corresponding tasks using large scale visual samples~\cite{li2017dual,gkioxari2015contextual,xu2015show}.
In contrast to their approaches, 
we propose to learn with multiple contextual regions guided by intention of human by utilizing the actor’s gaze cue.
With this, the framework is able to capture the dynamics of various HOIs with limited contextual region candidates.

In this work, we aim to tackle the challenge of accurately detecting and recognizing HOIs in social scene images.
We propose a human intention-driven HOI detection (iHOI) framework, consisting of an object detection module and three branches.
The first branch exploits individual features,
the second models differential human-object feature embeddings,
and the third leverages multiple gazed context regions in a weakly-supervised setting.
Human pose information has been incorporated into the feature spaces using the relative distances from body joints to the instances. 
The contributions of this work are summarized as follows:
\begin{enumerate}
	\item
	We have explored how to detect and recognize {\it what humans are doing in social scenes} by inferring {\it what they intended to do}.
	Specifically, 
	we provide a unique computational perspective to exploit human intention, commonly explored in cognitive studies,
	and propose a joint framework to effectively model gaze and body pose information to assist HOI detection.
	\item 
	We propose an effective hard negative sample mining strategy to address commonly observed mis-grouping problem in HOI detection, \ie~wrong object instance is assigned to the actor though with correct HOI category prediction.
	\item 
	We perform extensive experiments on two benchmark datasets with ablation studies, and show that iHOI outperforms the existing approaches. 
\end{enumerate}

The rest of the paper is as follows.
Section~\ref{sec:literature} reviews the related work.
Section~\ref{sec:method} delineates the details of the proposed method.
Section~\ref{sec:experiments} elaborates on the experiments and discusses the results. 
Section~\ref{sec:conclusion} concludes the paper.

\section{Related Work}
\label{sec:literature}
\noindent\textbf{Visual Relationship Detection.}
The inference of general visual relationships~\cite{lu2016visual,zhang2017visual,Krishna_CVPR_2018,li2017scene} has attracted increasing research interests.  
The types of visual relationships include verbs, preposition, spatial or comparative phrase.
Works have been attempted to refine visual relationships from vocabulary~\cite{lu2016visual}, 
embed object class probabilities to highlight semantics constraints~\cite{zhang2017visual},
disambiguate between entities of the same category with attention shifting conditioned on one another~\cite{Krishna_CVPR_2018}.
Visual relationships were also formalized as part of a graph representation for images called scene graphs~\cite{li2017scene}.
Our focus is related, but different.
We aim to explore direct interactions (actions) performed on objects, where human is the crucial indicator of the interactions. 

\noindent\textbf{HOI Understanding.}
Different from visual relationships detection task, 
which focuses on two arbitrary objects in the images, 
HOI recognition is a human-centric problem with fine-grained action categories. 
Earlier studies~\cite{gupta2009observing, yao2012recognizing, hu2013recognising} mainly focus on recognizing the interactions,
by joint modeling of body poses, spatial configuration, and functional compatibility in the images. 
In recent years, 
several human-centric image datasets have been developed to enable fine-grained exploration of HOI detection,
including V-COCO ({\it Verbs-COCO})~\cite{gupta2015visual}, HICO-DET ({\it Humans Interacting with Common Objects-DET})~\cite{chao2017learning}, and HCVRD ({\it Human-Centered Visual Relationship Detection})~\cite{zhuang2017care}.
In these datasets,  
the bounding boxes of each human actor and the interacting object are annotated, 
together with the corresponding interactions.  
Motivated by the success of deep learning,
especially Convolutional Neural Network for object detection and recognition,
several recent works have taken advantage of the detailed annotated datasets to improve HOI detection.
Works have been done for learning to detect HOIs with constraints from 
interacting object locations~\cite{gkioxari2017detecting,gupta2015visual}, 
pairwise spatial configuration~\cite{chao2017learning} 
to scene context of instances~\cite{Gao_BMVC_2018,Qi_ECCV_2018}.
Another stream of work addresses the long-tail HOI problem with compositional learning~\cite{Kato_ECCV_2018, shen2018scaling} and extra data supervision~\cite{zhuang2017care}.

In contrast to previous works treating humans and objects similarly, with no consideration that human behaviors are purposeful,
we argue that human intention drives interactions.
Therefore, in this work, we exploit the cues in an image that reflect an actor's intention,
and leverage such information for more effective HOI detection.

\begin{figure*}[!t]
	\centering
	\includegraphics[width=0.9\linewidth]{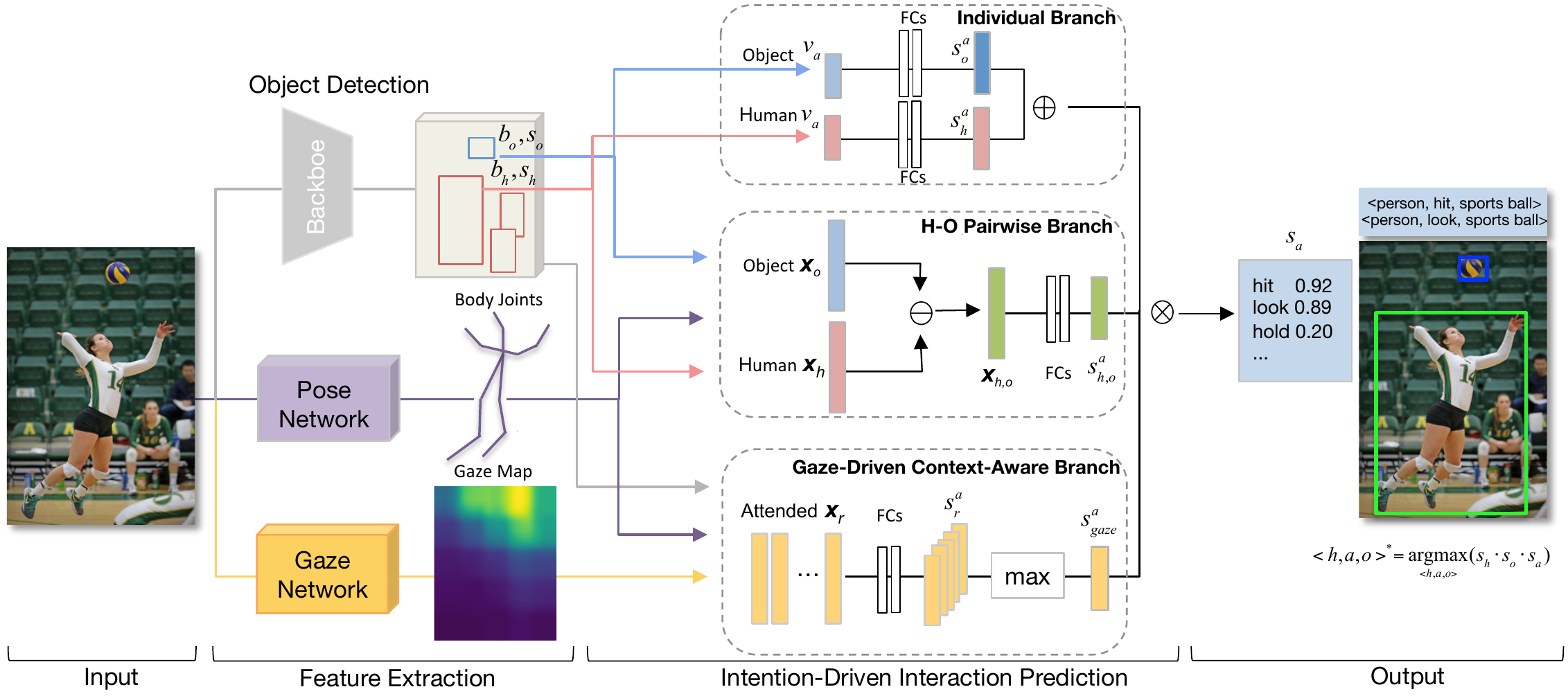}
	\caption
		{
		The proposed iHOI takes an image as input for feature extraction and human intention-driven interaction prediction, and outputs the detected triplets.
		Human intention is modelled as follows:
		1) the pose information is incorporated with the distances from body joints to the instance center;
		2) human gaze guides the attended context regions in a weakly-supervised setting.
		The feature spaces of $\Vec{x}_h$, $\Vec{x}_o$ and $\Vec{x}_r$ consist of: 
		class probabilities $\Vec{v}_c$,
		visual appearance $\Vec{v}_a$,
		relative locations $\Vec{v}_l$,
		and human pose information $\Vec{v}_p$.		
		Operations $\ominus$, $\oplus$ and $\otimes$ denote element-wise subtraction, summation and multiplication, respectively.
		} 
	\vspace{-3ex}
\label{fig:framework}
\end{figure*}

\noindent\textbf{Gaze in HOIs.}
Humans are the core element in HOIs.
Generally, 
an actor intends to leverage essential information in the scene to help performing the interaction.   
One important facet of intention is reflected by gaze, 
which explicitly shows the task-driven attention~\cite{van2010consciousness}.
Cognitive studies have reported that human often attends to the regions that provide significant information during an interaction~\cite{land2001ways}. 
Though this might not be true in some cases 
such as lifting a familiar object without specific attention, 
in general the fixated regions provide informative cues.

Inspired by the cognitive study findings, 
works have explored human gaze from computational perspectives.
Gaze locations have been predicted from first person view videos~\cite{fathi2012learning}, 
human head pose in multimodal videos~\cite{Mukherjee_2015_TMM},
and daily life images~\cite{recasens2015they}.
Prediction of human gaze direction has benefited saliency prediction~\cite{gorji2017attention}, 
and inspired prediction of shared attention in the crowds~\cite{Fan_2018_CVPR}.
Despite the efforts made in various tasks, 
the existing methods have yet to explore actor's gaze in the context of HOI detection. 
In this study, we explore the role of actor's gaze in guiding informative scene regions for HOIs.
\section{Proposed Method}
\label{sec:method}

This section presents the human intention-driven HOI detection (iHOI) framework, as shown in \fig~\ref{fig:framework}.
The task is formulated as follows: 
given a 2D image $\Vec{I}$ as the input, it aims to detect and recognize triplets of the form {\small $\langle \mathsf{human}, \mathsf{action}, \mathsf{object} \rangle$}.
We will first describe our model architecture, followed by the details of training that deals with mis-grouping problem (\ie~the class of a HOI is correctly predicted,
but a wrong object instance is assigned to the actor), and inference.  

\subsection{Model Architecture}
First, given an input image $\Vec{I}$,
we adopt Faster R-CNN~\cite{ren2015faster} from Detectron~\cite{Detectron2018} to detect all humans and objects, generating a set of detected bounding boxes $\Vec{b} = (b^1,...,b^m)$ where $m$ denotes the total number of detected instances.
The detected bounding boxes for a person and an object are denoted as $b_h$ and $b_o$, respectively. 
The corresponding confidence scores are denoted as $s_h$ and $s_o$, respectively.
Human body joints locations and gaze direction are obtained through transfer learning from other social-activity datasets~\cite{cao2017realtime, recasens2015they}, 
since our aim is to effectively model intention rather than extract features, and both experimental datasets lack the ground-truth.

Second, we predict the action prediction score $\Vec{s_a}$ for each candidate action $a \in \Vec{a}$ where $\Vec{a}$ is with dimension $A$ as the amount of all action classes, given each human-object bounding box pair ($b_h$ and $b_o$).
$\Vec{s_a}$ depends on:
(1) the action prediction confidence based on individual appearance of the person $\Vec{s_{h}^{a}}$ and the object $\Vec{s_{o}^{a}}$, 
(2) the action prediction confidence based on the human-object pairwise feature representations $\Vec{s_{h,o}^{a}}$, and
(3) the score prediction based on the gaze-driven contextual representations $\Vec{s_{gaze}^{a}}$.
Formally, given $b_h$ and $b_o$ with the respective $s_h$ and $s_o$, the action prediction score to maximize is as follows:
\begin{equation}
	\Vec{s_a} = \Vec{s_{h,o}^{a}} \cdot \Vec{s_{gaze}^{a}} \cdot (\Vec{s_{h}^{a}}+\Vec{s_{o}^{a}})
	\label{score}
\end{equation}

Multiplication in obtaining the final score aims to enlarge the contribution of each component (\ie~HO, context, H+O).
Sigmoid activation is used for multi-label action classification such that the predicted action classes do not compete. 
The training objective is to minimize the binary cross entropy loss between the ground-truth action labels $\Vec{y}$ and the predicted action scores $s_{ij} \in \Vec{s}$ with the following loss function:
\begin{equation}
    \mathcal{L}(\Vec{s}, \Vec{y}) = -\frac{1}{M}\sum_{j=1}^{M}\sum_{i=1}^{A} [y_{ij} \log\left(s_{ij}\right) + \left(1-y_{ij}\right)\log\left(1-s_{ij}\right)]  
\end{equation}
\vspace{-3ex}
\begin{equation}
	\mathcal{L}_{total} = \mathcal{L}(\Vec{s_{h,o}^{a}},\Vec{y})+\mathcal{L}(\Vec{s_{gaze}^{a}},\Vec{y})+ \mathcal{L}(\Vec{s_{h}^{a}},\Vec{y})+\mathcal{L}(\Vec{s_{o}^{a}},\Vec{y})
\end{equation}
where $\mathcal{L}$ is averaged on a batch of $M$ samples.
For the $j$-th prediction, $y_{ij}$ denotes the binary ground-truth action class for the $i$-th action,
and $s_{ij}$ denotes the prediction score for the $i$-th action after sigmoid activation.
$\Vec{s_{h,o}^{a}}$, $\Vec{s_{gaze}^{a}}$, $\Vec{s_{h}^{a}}$, and $\Vec{s_{o}^{a}}$ are the prediction confidence scores from the respective branch. 
During training, we train each branch independently because we want to maximize each branch's individual performance. During inference, we apply fusion from all branches so that we can obtain the optimal overall score.
An overview of the proposed model is shown in \fig~\ref{fig:framework}.

Next,
We describe each component of the architecture.
\subsubsection{Human-Object Pairwise Branch}
Humans and objects are paired according to the ground-truth during training.
Given the detected $b_h$ and $b_o$,
we aim to learn a pairwise feature embedding that can preserve their semantic interactions.
For example, the interaction of {\it a person riding a bike} can almost be described by the visual appearance of the pair, person on top of the object, and the estimated {\it bike} label.

Similar to the recent {\it VTransE}~\cite{zhang2017visual} for general visual relationships among objects, 
the feature space $\Vec{x}$ for each $b_h$ and $b_o$ contains visual appearance, relative spatial layout, and object semantic likelihood, 
referred as {\small $\Vec{v}_\alpha$}, {\small $\Vec{v}_l$} and {\small $\Vec{v}_c$}, respectively.
{\small $\Vec{v}_\alpha$} is a 2048-d vector, extracted from {\it fc7} layer in the object detector to capture the appearance of each $b_o$.
{\small $\Vec{v}_l$} is a 4-d vector consisted of {\small $\{l_x, l_y, l_w, l_h\}$}.
{\small $\{l_x, l_y\}$} specifies the bounding box coordinates distances, 
and {\small $\{l_w, l_h\}$} specifies the log-space height/width shift, 
all relative to a counterpart as parameterized in Faster R-CNN.
{\small $\Vec{v}_c$} is a 81-d vector of object classification scores over MS-COCO object categories, generated by the object detectors. 

In contrast to the general visual relationships, we extend the feature space with human pose information since our task is intrinsically human-centric.
Human pose bridges the human body with the interacting object. 
For example, the up-stretching arms, jumping posture and the relative distances to the ball possibly reveal that the person is {\it hitting a sports ball}. 
Since body pose ground-truth is not available, 
we use the pose estimation network in~\cite{cao2017realtime} to extract body joints locations for each human.
The output of the pose estimation network is the locations of 18 body joints.
We consider eight representative body joints\footnote{nose, neck, left and right shoulder, left and right elbow, left and right hip} that are more frequently detected,
which cover the head, upper and lower body.
For each joint $i \in {1,...,8}$, 
we calculate its distance from the center of $b_h$ and $b_o$ to get two distance vectors {\small $\{d_{xh}^i, d_{yh}^i\}$} and {\small $\{d_{xo}^i, d_{yo}^i\}$}, 
where {\small $d_{xh}^i$} denotes its distance from $b_h$ center along x-axis, and {\small $d_{yo}^i$} denotes the distance from $b_o$ center along y-axis.
Since human-object pairs have different scales, we normalize the distances {\it w.r.t.} the width of $b_h$.
We concatenate the normalized distance vectors for all eight joints to get two 16-d vectors {\small $\Vec{v}_p^h = \{d_{xh}^i, d_{yh}^i|i = 1,...,8\}$} and {\small $\Vec{v}_p^o = \{d_{xo}^i, d_{yo}^i|i = 1,...,8\}$} that encode the pose information.
In cases where not all eight joints are detected, we set {\small $\Vec{v}_p$} to be zeros.
An alternative way of implementing the pose information have been experimented, as shown in Section~\ref{sec:experiments}.

The above-mentioned features are concatenated to form the feature spaces for human 
{\small $\Vec{x}_h=\{\Vec{v}_c^h, \Vec{v}_\alpha^h, \Vec{v}_l^h, \Vec{v}_p^h\}$}
and object {\small $\Vec{x}_o=\{\Vec{v}_c^o, \Vec{v}_\alpha^o, \Vec{v}_l^o, \Vec{v}_p^o\}$}.
Following \cite{zhang2017visual, Bordes_2013_NIPS},
we calculate the pairwise feature embedding as
\begin{equation}
\Vec{x}_{pair} = \Vec{x}_{h} \ominus \Vec{x}_{o}
\end{equation}
The differential embedding tries to represent the pairwise relation as a translation in the embedding space~\cite{Bordes_2013_NIPS}: $\mathsf{human}+\mathsf{interaction}=\mathsf{object}$.
Pairwise feature summation has also been experimented but shown less effectiveness.
The pairwise embedding is passed through a fully-connected layer to produce the pairwise action scores $\Vec{s_{h,o}^{a}}$. 

\subsubsection{Individual Human and Object Branch}
Individual classification based on visual appearance of human and object has been commonly used and approved to be effective~\cite{chao2017learning,Gao_BMVC_2018}.
For example, the shape and size of an elephant can help to infer action as ``riding'' rather than ``holding''.
We leverage a human stream and an object stream to predict interactions individually.
Within each stream, two fully connected layers (FCs), each followed by a dropout layer, are adopted.
The inputs are the visual appearance $\Vec{v}_\alpha$ extracted for $b_h$ and $b_o$, respectively.
Following the late fusion strategy, each stream performs action prediction first, then two predicted scores based on appearance from human $\Vec{s_{h}^{a}}$ and object $\Vec{s_{o}^{a}}$ are fused by element-wise summation with equal proportion.

\subsubsection{Gaze-Driven Context-Aware Branch} 
\label{subsubsec:gaze}

We observe that the regions where an actor is fixating often contain useful information for the interactions.
For example, when the person intends to pour water to a cup, he normally {\it fixates around the cup while holding it}.
Therefore, we exploit the fixated contextual information to help recognizing the actor's action.
In particular, we use human gaze as a guidance to leverage the fixated scene regions.
The gazed location is predicted with a pretrained two-pathway model proposed in~\cite{recasens2015they}.
The prediction is reasonable only if the visualized region lies in the line defined by the person’s eye location and head orientation. 
Since there is no gaze ground-truth, we did manual qualitative check. 
Among 100 random gaze predictions, there were 64 correct predictions, 24 false predictions and 12 predictions that cannot be decided where the actor’s eyes are invisible in images, or the actor faces to the frontal direction to the camera.
The gaze prediction model takes the image $\Vec{I}$ and the central human eye position (calculated from the pose estimation network) as input, 
and outputs a probability density map $\Vec{G}$ for the fixation location.

For each human in the image, we select five regions from the candidates $\Vec{b} = (b^1,...,b^m)$ generated by the object detectors, which have higher likelihood of being fixated on.
Specifically, for each candidate region $b \in \Vec{b}$, we assign a gaze weight ${g}_{b}$ to it,
where ${g}_{b}$ is obtained by summing up the values of $\Vec{G}$ in $b$ and then normalized by the area of $b$:
\begin{equation}
{g}_{b} = \frac{\sum_{{x,y} \in b}{\Vec{G}_{x,y}}}{{area_b}}, b \in \Vec{b}
\end{equation}

Then we select the top-5 regions $\Vec{r} = (r^1,...,r^5)$ that have the largest ${g}_{b}$.
We have experimented with different numbers of candidate regions from one to all of the detected objects.
Using top-5 candidate regions guided by gaze achieves plateau performance, which suggests that five candidates are sufficient to capture informative cues.
With the same definition of feature representations for human and object instances, 
for each selected region $r \in \Vec{r}$, we first get its corresponding feature vector {\small $\Vec{x}_r=\{\Vec{v}^r_c, \Vec{v}^r_a, \Vec{v}^r_l, \Vec{v}^r_p\}$}.
We then pass it through a FC layer to acquire the action scores $\Vec{s_{r}^{a}}$ for each of the regions $\Vec{r}$.
We compute the prediction score for this branch as follows:
\begin{equation}
\Vec{s_{gaze}^{a}}=\max(\Vec{s_{r}^{a}}), r \in \Vec{r}
\end{equation}
$\max(\cdot)$ is used because generally there is only one region an actor can fixate on.
The most informative region among the gazed candidates can be discovered in a weakly-supervised manner.
Note that if the gaze of the actor cannot be predicted (\ie~the eyes are invisible in images, or the actor faces to the frontal direction to the camera),
we set $\Vec{x}_{r}=0$.

In contrast to a recent work~\cite{Wei_CVPR_2018} directly leveraging the fixated patch, 
learning with multiple gazed regions makes it robust to the inaccurate gaze predictions 
and dynamic to different HOIs with the same gaze direction, 
\ie~it can find the most informative region among a reasonable amount of guided candidate regions for the corresponding HOI.

\vspace{-1ex}
\subsection{Hard Negative Triplet Mining}
We observe that mis-grouping is a common category of false positive HOI detection~\cite{gupta2015visual}.
Mis-grouping refers to cases where the class of a HOI is correctly predicted,
but a wrong object instance is assigned to the actor
(\eg~a person is {\it cutting} another person's cake). 
We argue that such negative HOI triplets are more difficult for a model to reject in the tasks requiring pairing proposals, due to less discriminative patterns,
compared with other negative triplets of inaccurate localization and false classification in~\cite{chao2017learning}.

We propose a simple yet effective method to mine for those hard negative triplets.
For each image, we deliberately mis-group non-interacting human-object pairs from the annotation,
and set their action labels as negative, \ie~all zeros.
These human-object pairs together with negative labels form the negative triplets.
We adopt the image-centric training strategy~\cite{girshick2015fast}, 
where each mini-batch of HOI triplets arises one image.

\vspace{-1ex}
\subsection{Inference}
\label{subsubsec:inference}

During inference, we aim to calculate the HOI score $s_{h,o,a}$ for a triplet {\small $\langle \mathsf{human}, \mathsf{action}, \mathsf{object} \rangle$}.
Given the detection score for human $s_{h}$, object $s_{o}$ (generated from the object detection module directly), and the largest action confidence $s_a$ among the scores for all actions $\Vec{s_a}$ (\eqn~\ref{score}), $s_{h,o,a}$ is obtained by $s_h \cdot s_o \cdot s_a$.
For some of the action classes that do not involve any objects (\eg~run, walk, smile), we use the HOI score $s_{h,o,a}$ based on only the human information: $s_{h,o,a} = s_h \cdot s_{gaze}^{a} \cdot s_{h}^{a}$.

To predict HOIs in an image, we must compute the scores for all detected triplets.
However, scoring every potential triplet is almost intractable in practice, calling for high-recall human-object proposals.
To solve this, we leverage the predefined relevant object categories $c \in C$ for each action~\cite{gupta2015visual} as a prior knowledge, which is extracted from the HOI ground-truth in the corresponding dataset.
For instance, {\it sports ball} is relevant to the action {\it kick} but {\it book} is not relevant. 
Unlike pairing human and objects according to the ground-truth during training,
we filter out the detected objects irrelevant to the action for each human-action pair during inference/testing.
We then select the object that maximizes the triplet score $s_{h,o,a}$ within each relevant category to form the triplet.
Note that for HICO-DET, there exist many samples of human interacting with multiple objects of the same category (\eg~{\it a person is herding multiple cows}), therefore we retain at most 10 objects sorting by $s_{h,o,a}$ for each human-action pair.

With objects selected for each human and action, we have triplets of {\small $\langle \mathsf{human}, \mathsf{action}, \mathsf{object} \rangle$}.
The bounding boxes of the human-object pairs, along with their respective HOI triplet score $s_{h,o,a}$, are the final outputs of our model.
 
\section{Experiments}
\label{sec:experiments}
In this section, 
we first introduce the datasets and evaluation metrics, then describe the implementation details.
Next,
we compare our proposed iHOI framework to the state-of-the-art methods quantitatively.
Finally,
we conduct ablation studies to examine the effect of each proposed component,
and qualitatively evaluate the results.

\subsection{Datasets and Evaluation Metrics}
\label{sec:experiments_data}

We adopt two HOI detection datasets V-COCO~\cite{gupta2015visual} and HICO-DET~\cite{chao2017learning}.
The other datasets~\cite{chao2015hico,zhuang2017care} either is not in the context of detection task or contains general human-object predicates that are out of our exploration range.

\noindent\textbf{V-COCO Dataset~\cite{gupta2015visual}} 
is a subset of MS-COCO~\cite{lin2014microsoft}, with 5,400 images in the trainval (training plus validation) set and 4,946 images in the test set.
It is annotated with 26 common action classes (five of them have no interacting objects), and the bounding boxes for human and interacting objects.
In particular, 
three actions (\ie~cut, hit, eat) are annotated with two types of targets (\ie~instrument and direct object). 

\noindent\textbf{HICO-DET Dataset~\cite{chao2017learning}}
contains 38,118 images in the training set and 9,658 in the test set. 
It is annotated with 600 types of interactions:
80 object categories as in MS-COCO and 117 verbs.

\noindent\textbf{Evaluation Metrics.}
We evaluate mean Average Precision (mAP) for both datasets.
Formally, a triplet of {\small $\langle \mathsf{human}, \mathsf{action}, \mathsf{object} \rangle$} is considered as a true positive if: 
1) both the predicted human and object bounding boxes have IoUs $\geq 0.5$ with the ground-truth, and
2) the predicted and ground-truth HOI classes match. 
The definition of true positive is identical except that HICO-DET considers the specific object categories,
while V-COCO considers rough object types, namely {\it instrument} and {\it direct object}, as in the standard evaluation metric.
For V-COCO, we evaluate $mAP_{role}$ following~\cite{gkioxari2017detecting}. 
For HICO-DET, we follow the evaluation setting~\cite{chao2017learning} with objects unknown in advance: Full (600 HOIs), Rare (138 HOIs), and Non-Rare (462 HOIs). 

\begin{figure*}[!t]
	\centering
	\includegraphics[width=1.0\linewidth,trim={0 0 0 12},clip]{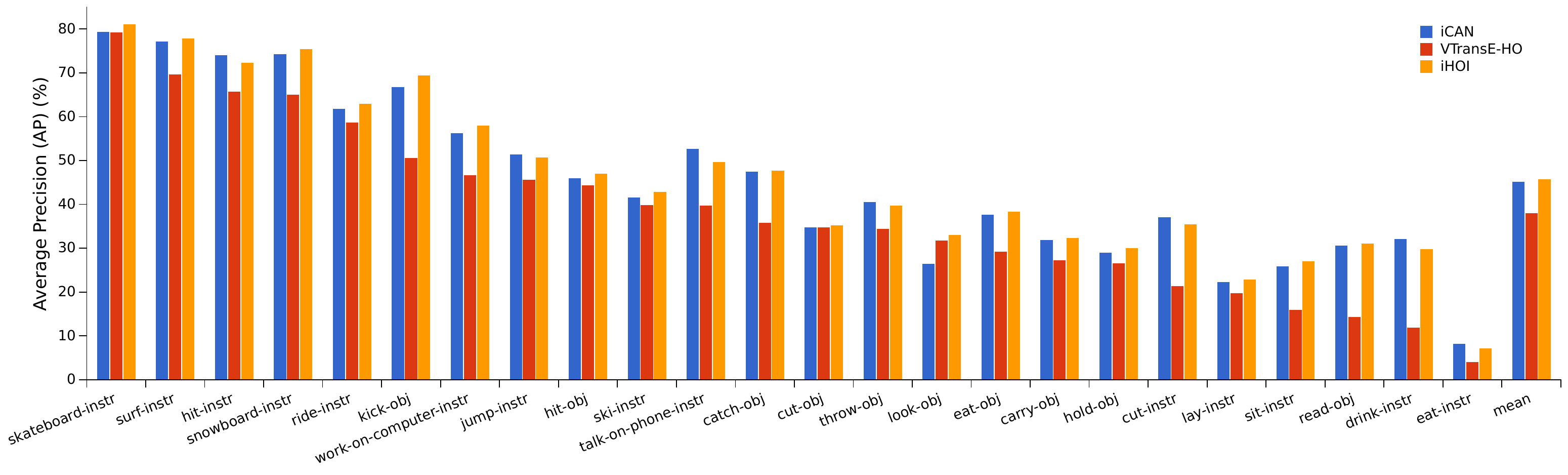}
	\vspace{-4ex}
	\caption
	{
		Per-action mAP (\%) of the triplets (${mAP}_{role}$) on V-COCO test set. 
		We show two main baselines and our framework for each of the actions with interacting objects. 
		There are 26 actions defined in~\cite{gupta2015visual}, 
		and five of them are defined without objects.  
		We list the detailed results for 21 actions with all possible types of interacting object.
	}
	\vspace{-1.5ex} 
	\label{fig:vcoco_detail}
\end{figure*}

\subsection{Implementation Details}
Our implementation is based on Faster R-CNN~\cite{ren2015faster} with a Feature Pyramid Network (FPN)~\cite{lin2017feature} backbone built on ResNet-50~\cite{he2016deep}, from Detectron~\cite{Detectron2018}.
The weights are pretrained on MS-COCO dataset.
Following the same thresholds~\cite{Gao_BMVC_2018}, human and object bounding boxes with detection scores above 0.8 and 0.4 are kept, respectively.
The object detection backbone is kept frozen during training.
We follow the image-centric training strategy with mini-batch size set to 32.
We adopt Stochastic gradient descent (SGD) to train the model for 20 epochs with a learning rate of 0.001, a weight decay of 0.0005, and a momentum of 0.9.

\begin{table}[!t]
	\centering
	\caption
		{
		Comparisons with the state-of-the-art methods and variants of iHOI on V-COCO dataset.
		mAP (\%) equals to $mAP_{role}$ as in the standard evaluation metric.
		}
	\label{tb:vcoco}
 	\vspace{-1ex}
	\begin{tabular}{l|c}
		\toprule	
		Methods\hspace{30ex}                                            		 & mAP (\%) \\	
		\hhline{==}
		VSRL~\cite{gupta2015visual}                                     		 & 31.80 \\
		InteractNet~\cite{gkioxari2017detecting}                         & 40.00 \\
		GPNN~\cite{Qi_ECCV_2018}				  							& 44.00 \\
		iCAN~\cite{Gao_BMVC_2018}				  						  & 45.30  \\
		\midrule
		VTransE-HO														    & 38.09 \\
		\CellY{(a) w/ pose locations}                                   	 & \CellY{38.62} \\
		\CellY{(b) w/ P (pose distances)}                                  & \CellY{38.89} \\
	    \CellB{(c) w/ sorted $r$}                                       	& \CellB{40.85} \\
		\CellB{(d) w/ G (gazed $r$)}                                    	 & \CellB{41.62} \\
		\CellG{(e) w/ P+G}                                              	& \CellG{42.37} \\
		\CellG{(f) w/ P+G+an alternative mining~\cite{chao2017learning}} & \CellG{44.61} \\
		\midrule
		{\bf iHOI}                                  & \textbf{45.79}   \\
		\bottomrule
	\end{tabular}
 	\vspace{-2ex}
\end{table}

The aim of this work is to effectively model human intention into HOI detection framework rather than extract features.
Therefore, human gaze and pose information are transferred from other social-activity datasets~\cite{cao2017realtime, recasens2015they}.
Our framework could be further trained in an end-to-end manner if the human gaze and body pose annotations are available.
\vspace{-2ex}

\subsection{Comparing with Existing Methods}
We compare our proposed iHOI with existing methods for HOI detection. 
Specifically, 
we compare with four methods~\cite{gkioxari2017detecting,gupta2015visual,Qi_ECCV_2018,Gao_BMVC_2018} on V-COCO, 
and five methods~\cite{gkioxari2017detecting,chao2017learning,Gao_BMVC_2018,Qi_ECCV_2018,shen2018scaling} on HICO-DET.


In general, 
\tab~\ref{tb:vcoco} and \tab~\ref{tb:hico_det} show that iHOI outperforms other methods.
HICO-DET is generally observed with lower mAP because it contains more fine-grained HOI categories with severe long-tail problem, and is evaluated with specific object categories rather than the two rough types of objects in V-COCO.
Our iHOI outperforms the best performing methods (\ie~iCAN and GPNN, respectively) with improvements of +0.49 on V-COCO and +0.28 on HICO-DET full test set.
Existing methods mainly rely on the representations of human-object appearance and spatial relationships through ConvNets or Graph Neural Networks.
However, 
some complicated interactions are very fine-grained, which make it hard to distinguish only by appearance and relative locations.
On the other hand,
the proposed iHOI jointly takes advantages of the gazed scene context and subtle differences of the body movements.
A discriminative pattern between the positive and hard negative samples is also learnt.
Thus it achieves better overall performance. 

\begin{table}[!t]
	\centering
	\caption
		{
		Comparisons with the state-of-the-art methods and variants of iHOI on HICO-DET dataset.
		Results are reported with mean Average Precision (mAP) (\%).
		}
	\label{tb:hico_det}
	\vspace{-1ex}
	\resizebox{0.97\linewidth}{!}{$
	\begin{tabular}{l|c c c}
		\toprule	
			Methods\hspace{20ex}    & \hspace{2ex}Full\hspace{2ex} & \hspace{2ex}Rare\hspace{2ex} & Non-Rare \\ 
		\hhline{====}
		Shen~\etal~\cite{shen2018scaling}        & 6.46	& 4.24 & 7.12 \\
		HO-RCNN~\cite{chao2017learning}          & 7.81	& 5.37 & 8.54 \\
		InteractNet~\cite{gkioxari2017detecting} & 9.94 & 7.16 & 10.77 \\
		GPNN~\cite{Qi_ECCV_2018}				 	& 13.11 & 9.34 & 14.23 \\	
		iCAN~\cite{Gao_BMVC_2018}   			& 12.80 & 8.53 & 14.07 \\
		\midrule
		VTransE-HO 	&    10.98 & 8.15	  &      11.82 \\
		\CellY{(a) w/ pose locations}		 & \CellY{11.21} & \CellY{8.17} & \CellY{12.12} \\
		\CellY{(b) w/ P (pose distances)}	 & \CellY{11.51} & \CellY{8.21} & \CellY{12.49} \\
		\CellB{(c) w/ sorted $r$}            & \CellB{11.89} & \CellB{8.73} & \CellB{12.83} \\
		\CellB{(d) w/ G (gazed $r$)}         & \CellB{12.03} & \CellB{8.89} & \CellB{12.97} \\
		\CellG{(e) w/ P+G}                   & \CellG{12.40} & \CellG{8.92} & \CellG{13.44} \\
		\CellG{(f) w/ P+G+an alternative mining~\cite{chao2017learning}} & \CellG{12.96} & \CellG{9.27} & \CellG{14.06} \\
		\midrule
	    {\bf iHOI}                 & {\bf 13.39} & {\bf 9.51} & {\bf 14.55} \\
		\bottomrule 
	\end{tabular}
	$}
 	\vspace{-2ex}
\end{table}

\begin{figure*}[!t]
	\centering
	\includegraphics[width=1.0\linewidth]{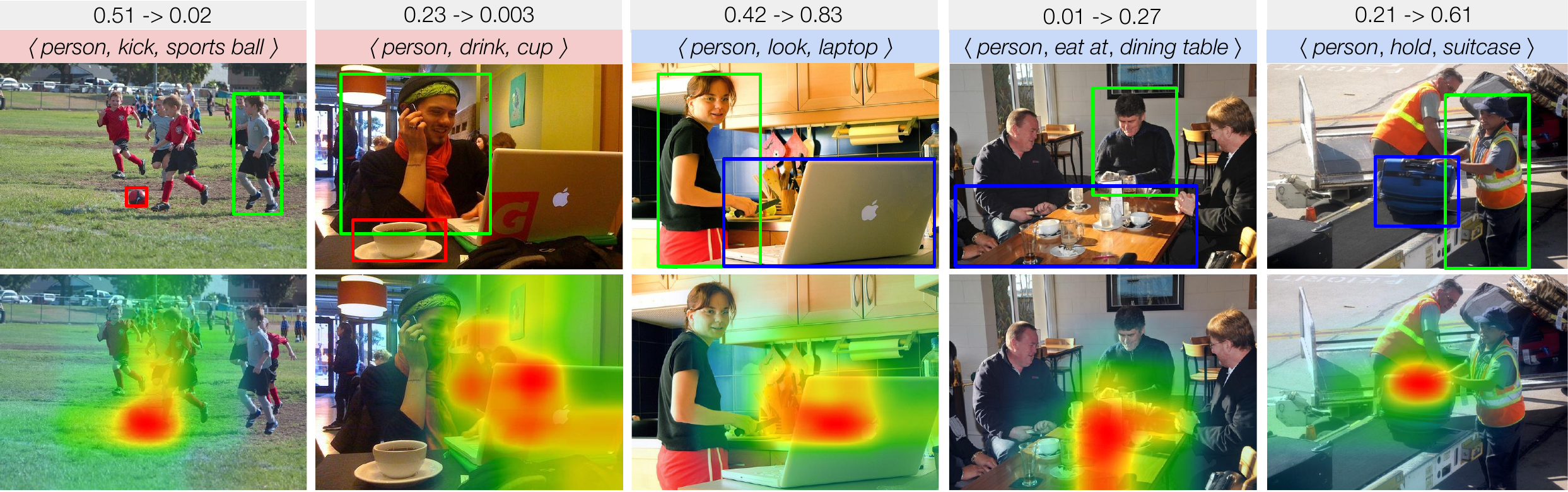}
	\caption
		{
		The effect of human intention on both datasets.
		HOI predictions together with the triplet scores (with \textcolor{gray}{gray} headings) are shown. 
		After leveraging intention ({\it VTransE-HO w/ P+G} vs. {\it VTransE-HO}), we show the change in triplet scores.
		Using intention suppresses the false prediction scores (\ie~column 1--2 with \textcolor{red}{red} headings),
		whereas improves the correct ones (\ie~column 3--5 with \textcolor{blue}{blue} headings). 
		The corresponding gaze density heatmaps intuitively demonstrate that fixated regions are informative of HOIs.
		The pose information is not plotted.
		} 
		\vspace{-3.5ex}
\label{fig:intention}
\end{figure*}

To study the effectiveness on various interaction classes,
we analyze the mAP for each action-target type defined in V-COCO. 
\fig~\ref{fig:vcoco_detail} shows the detailed results of iCAN, the base framework VTransE-HO, and our proposed iHOI.
VTransE-HO consists of action predictions from individual human and object appearance, as well as the pairwise feature embeddings of {\small $\{\Vec{v}_c, \Vec{v}_\alpha, \Vec{v}_l\}$}~\cite{zhang2017visual}.
We observe consistent actions with leading mAP, such as {\it skateboard, surf}.
Our proposed framework improves the performance on all action-target categories compared to VTransE-HO. 
The actions with the largest improvement are those closely related with human intention such as {\it read, work-on-computer}, 
those requiring the contextual information such as {\it drink},
and those likely to be mis-grouped such as {\it kick, snowboard}. 
Unlike modeling the visual scene context with convolutional kernel in iCAN~\cite{Gao_BMVC_2018}, our work attempts to extract informative scene regions with human-centric cues. 
Comparing the proposed iHOI with the best performing iCAN shows that iHOI can achieve overall better performance on most action-target categories,
whereas showing slight worse performance on a small proportion of the categories such as {\it hit-instr, talk-on-phone-instr, cut-instr}.
We observe that iHOI performs worse mostly on actions with small objects, possibly due to inaccurate object detection.

\vspace{-2ex}
\subsection{Ablation Studies}
In this section, 
we examine the impact of each proposed component with the following iHOI variants upon the base framework VTransE-HO, shown in \tab~\ref{tb:vcoco} and \tab~\ref{tb:hico_det}:
\begin{enumerate}[label=(\alph*)]
	\item {\bf w/ pose locations}: The relative locations of body joints w.r.t the image size are used to compute an additional set of action scores.
	\item {\bf w/ P (pose distances)}: The relative distances from body joints to the instance are concatenated into the respective human and object feature spaces, as in iHOI. 
	\item {\bf w/ sorted $r$}: An additional context-aware branch is implemented, and the top-5 scene regions are selected by detection scores (w/o pose).
	\item {\bf w/ G (gazed $r$)}: An additional gaze-driven context-aware branch is implemented (w/o pose).
	\item {\bf w/ P+G}: Body joints distances and gaze information are incorporated into the two-branch model, equivalent to the proposed iHOI without hard negative triplet mining.
	\item {\bf w/ P+G+an alternative mining~\cite{chao2017learning}}: A general mining method~\cite{chao2017learning} is used in addition to (e), to compare with our proposed mining strategy.
\end{enumerate}

\begin{figure}[!t]
	\centering
	\includegraphics[width=1.0\linewidth]{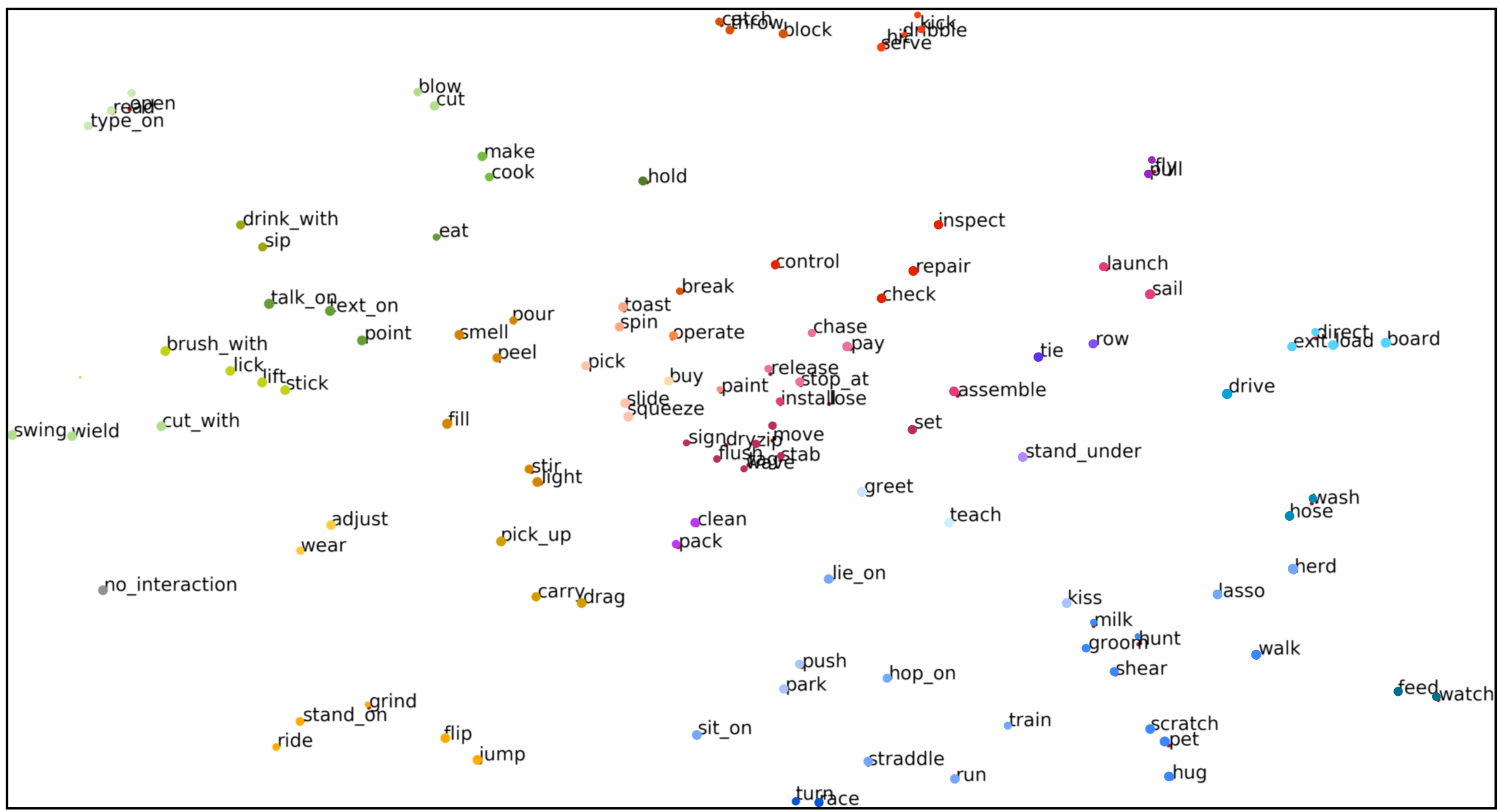}
	\caption
		{
        Visualization of the 117 action model parameters of the human-object pairwise feature embedding using t-SNE~\cite{maaten2008visualizing}.
		} 
	\vspace{-4.5ex}
\label{fig:tsne}
\end{figure}

The reimplemented base framework VTransE-HO achieves solid performance on both datasets, 
based on individual appearance and pairwise differential feature embedding~\cite{zhang2017visual}.
To gain insight into the learned pairwise embedding, we explore whether the embedding space has captured certain clustering properties in \fig~\ref{fig:tsne} with t-SNE visualization~\cite{maaten2008visualizing}.
By inspecting the semantic affinities of the embeddings, 
we can observe that the learned embedding space yields a semantically reasonable projection of data, shown with different colors.
Our iHOI achieves gains in mAP of +7.70 on V-COCO and +2.41 on HICO-DET, which are relative improvements of 20.22\% and 21.95\% over VTransE-HO. 
We analyze the effect of each component as follows. 

\begin{figure*}[!t]
	\centering
 	\includegraphics[width=1.0\linewidth]{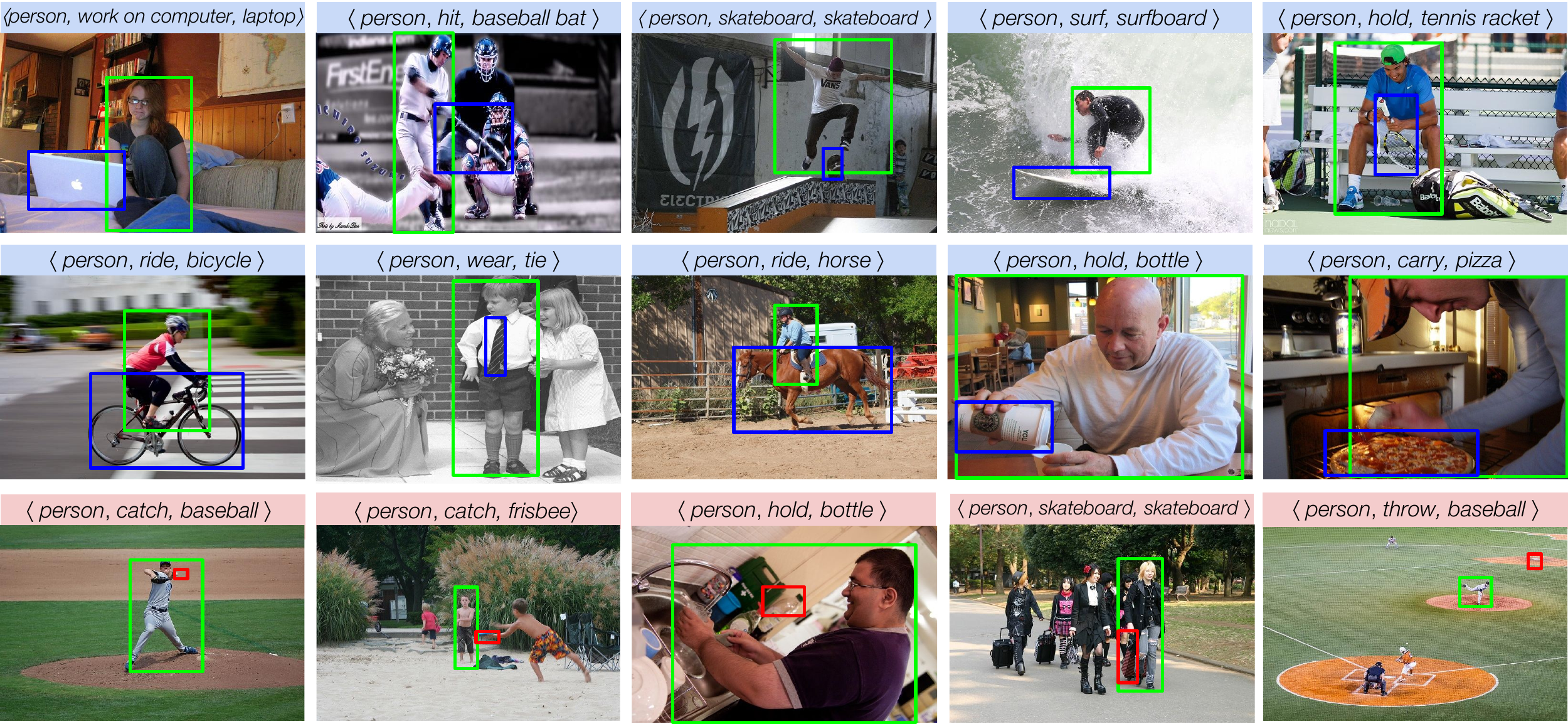}
	\caption
		{
		Samples of human-object interactions detected by our proposed method. 
		Each image displays top-1 {\small $\langle \mathsf{human}, \mathsf{action}, \mathsf{object} \rangle$} triplet.
		The first two rows present correct detections, 
		and the last row presents false positives.		
		} 
	 	\vspace{-3ex}
\label{fig:tp_fp}
\end{figure*}
\begin{figure}[!h]
	\centering
	\includegraphics[width=1.0\linewidth]{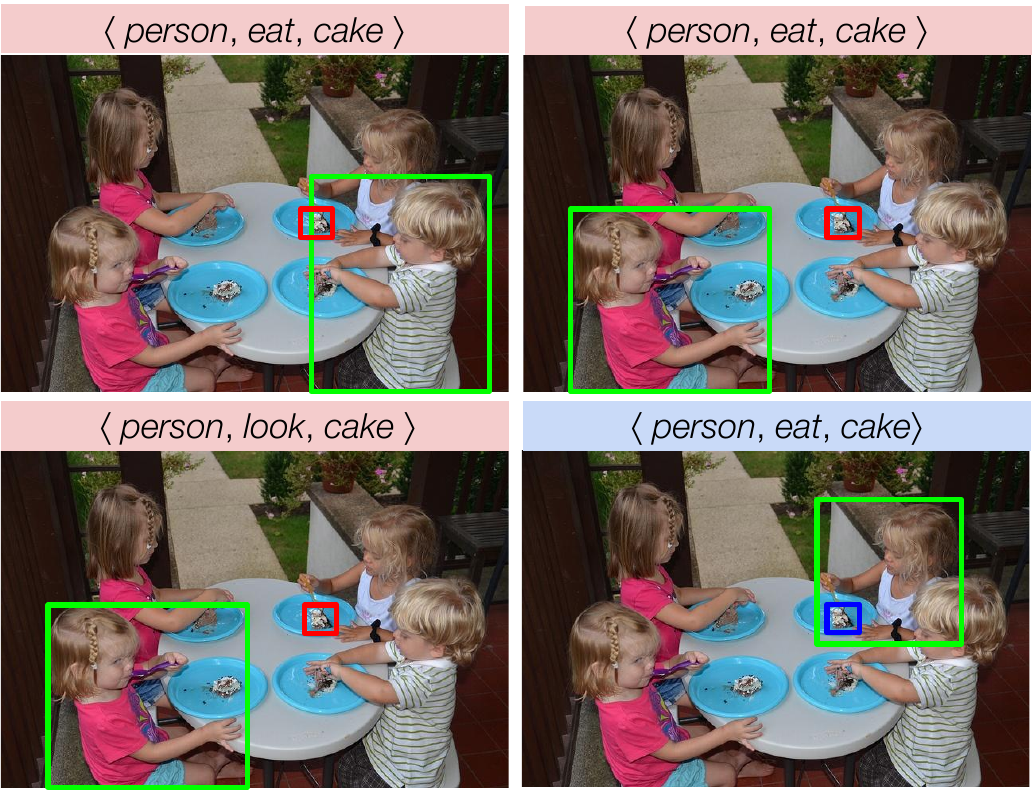}
	\caption
		{
		Detections with triplet score larger than 0.7 are displayed.
		Without the proposed negative triplet mining, the model gives all four predictions, in which three of them (\ie~with \textcolor{red}{red} headings) are mis-grouped.
		Model with negative triplet mining reduces the prediction to only the correct triplet (\ie~the bottom right with \textcolor{blue}{blue} heading). 
		} 
	\vspace{-4ex} 
\label{fig:mining}
\end{figure}

\subsubsection{Gazed Context}
Human gaze explicitly conveys his/her intention,
which drives the attended scene regions in the intention-driven branch, forming a key component of our framework.
The ablation results are colored with blue.
(c) leverages the scene regions sorted by detection scores without gaze guidance,
which improves upon the base framework VTransE-HO with +2.76 and +0.91 on V-COCO and HICO-DET, respectively. 
The improvements indicate that scene regions are informative for HOI detection.
By utilizing the actor's fixated regions guided by gaze,
(d) further achieves improvements of +0.77 and +0.14 compared to (c).
This demonstrates that the actor's fixated regions can reasonably provide information in detecting HOIs
even with some ambiguous gaze predictions.
The effectiveness of using the gazed context is also demonstrated by (e) vs. (b).

The optimal value for the hyper-parameter (\ie~number of candidate) depends on the image context and the gaze prediction accuracy. Using fewer candidates makes the model more sensitive to gaze prediction whereas using more candidates increases its robustness to gaze prediction at the cost of introducing noisy context. In our work, we empirically find that using the top-5 regions achieve the best performance.

\subsubsection{Human Pose}
Human pose implicitly conveys his/her intention,
which bridges the action with the interacting object.
Comparing (b) to VTransE-HO,
incorporating joints distance information achieves +0.80 and +0.53 in mAP on the two datasets.
Comparing method (e) to (d),
the improvements are +0.75 and +0.37, respectively.
It shows that HOI recognition is likely to be benefited from capturing the subtle differences of body movements.
Yet the performance improvement is slight, possibly because
the pose prediction could be inaccurate due to scale variation, crowding, occlusion.

An alternative implementation of human pose information has also been conducted, 
which directly incorporating the body joints coordinates, shown in (a) with yellow color.
The advantage of (b) over (a) demonstrates the efficacy of the proposed implementation of pose information.
In particular,
iHOI can capture the spatial differences of movements relative to the human and object. 

\subsubsection{Modeling Human Intention}
Human intention can be jointly modeled using both gaze (G) and pose (P).
Comparing (e) to VTransE-HO,
considering both gaze and pose achieves +4.28 and +1.42 in mAP for the two datasets.

Qualitatively,
\fig~\ref{fig:intention} shows five HOI predictions with notable changes in the scores after joint modeling intention using gaze and pose. 
The false triplet predictions (\ie~with red headings) are suppressed by incorporating human intention.
For example,
in the first image,
it is unlikely that the detected boy is {\it kicking the sports ball} due to
the large distances between his body joints to the target ball, as well as there is another boy nearer to the ball with a kicking pose.  
In the second image,
the score of {\it drinking with cup} is significantly decreased when the model learns that the person is looking at a laptop.
Meanwhile, leveraging human intention increases the confidence of correct HOI predictions, shown by examples with blue headings.
It indicates that human intention can reasonably help by leveraging the gazed context and the spatial differences of pose.

\subsubsection{Hard Negative Triplet Mining}
The ablation results for an alternative negative mining and the proposed one are colored with green.
Without the proposed mining strategy, shown in (e), mAP is decreased by -3.42 and -0.99 on the two datasets.
This demonstrates that the examined hard negative samples are essential for the model to learn a more discriminative pattern.
Our method specifically targets the hard triplets that are likely to be mis-grouped,
therefore outperforms the general negative mining of inaccurate localization and false classification~\cite{chao2017learning}, shown in (f).
Our proposed mining method can be applied to other tasks that require pairing of proposals.

\fig~\ref{fig:mining} shows the effectiveness of the proposed negative triplet mining strategy for HOI detection.
If no negative sampling is used,
there exists interaction hallucination (\ie~{\it eat the other person's cake}).
Model trained with the proposed strategy manages to reject the mis-grouped pairs and only predicts the correct triplet
(\ie~the bottom right prediction).

\subsubsection{Qualitative Examples}
\fig~\ref{fig:tp_fp} shows the examples of HOI detections generated with the proposed iHOI framework, including correct predictions and false positives.
The incorrect detections can be caused by confusing actions (\eg~{\it catch} and {\it throw} sports ball), inaccurate object detections (\eg~object detected on the background, false object classification or localization), and incomplete HOI annotations.

 
\section{Conclusion}
\label{sec:conclusion}

In this work, we introduce a human intention-driven framework, namely iHOI, to detect human-object interactions in social scene images. 
We provide a unique computational perspective to explore the role of human intention, 
\ie~iHOI jointly models the actor's attended contextual regions, and the differences of body movements.
In addition, 
we propose an effective hard negative triplet mining strategy to address the mis-grouping problem.
We perform extensive experiments on two benchmark datasets, 
and validates the efficacy of the proposed components of iHOI.
Specifically,
learning with multiple gaze guided scene regions makes the detection dynamic to various interactions.
For future work,
gaze prediction on small objects could be explored,
which the current model is weak at.
Another direction could be studying human intention for HOI detection in videos,
where intention is conveyed through spatial-temporal data.

\balance
\bibliographystyle{IEEEtran}
\bibliography{reference}


\end{document}